\documentclass[10pt,twocolumn,letterpaper]{article}

\usepackage{iccv}              

\definecolor{iccvblue}{rgb}{0.21,0.49,0.74}
\usepackage[pagebackref,breaklinks,colorlinks,allcolors=iccvblue]{hyperref}
\usepackage{soul} 
\usepackage{multirow}

\def\paperID{5906}
\def\confName{ICCV}
\def\confYear{2025}

\title{Aligning Constraint Generation with Design Intent in Parametric CAD}

\author{
Evan Casey \and
Tianyu Zhang \and
Shu Ishida \and
William P. McCarthy \and
John R. Thompson \and
Amir Khasahmadi \and 
Joseph G. Lambourne \and
Pradeep Kumar Jayaraman \and
Karl D.D. Willis \and \\ 
Autodesk Research 
}

\begin{document}

\twocolumn[{%
\renewcommand\twocolumn[1][]{#1}%
\maketitle



\vspace{-1em}
\includegraphics[width=\linewidth]{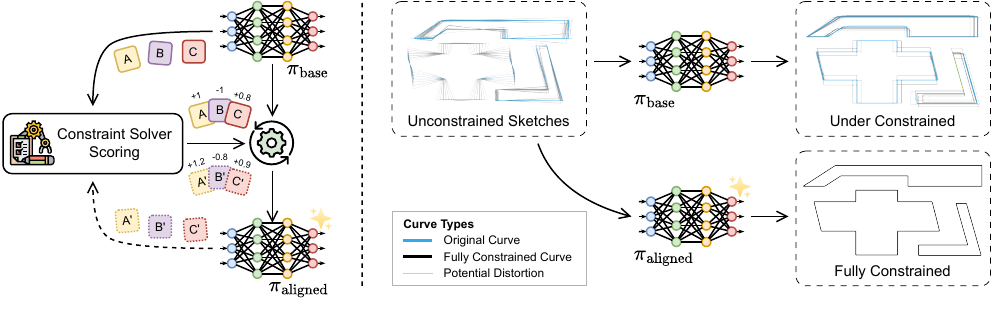}

\centering
\captionof{figure}{\textbf{Left:} a constraint solver is used to score model generated constraints $\texttt{A}, \texttt{B}, \texttt{C} \sim \pi_{\text{base}}$ (and $ \texttt{A}’, \texttt{B}’, \texttt{C}’ \sim \pi_{\text{aligned}}$). Starting with the base model $\pi_{\text{base}}$, we post-train an aligned model $\pi_{\text{aligned}}$ from this feedback.
\textbf{Right:} Blue lines show original primitives and gray lines show geometric distortion when dimensions vary. The aligned model $\pi_{\text{aligned}}$ produces fully-constrained sketches that preserve relative geometric relationships, whereas the base model $\pi_{\text{base}}$ produces under-constrained sketches that may distort the geometry in unintended ways.}

\vspace{1em}
}]

\begin{abstract} 
We adapt alignment techniques from reasoning LLMs to the task of generating engineering sketch constraints found in computer-aided design (CAD) models. Engineering sketches consist of geometric primitives (e.g. points, lines) connected by constraints (e.g. perpendicular, tangent) that define the relationships between them. For a design to be easily editable, the constraints must effectively capture design intent, ensuring the geometry updates predictably when parameters change. Although current approaches can generate CAD designs, an open challenge remains to align model outputs with design intent, we label this problem `design alignment'. A critical first step towards aligning generative CAD models is to generate constraints which fully-constrain all geometric primitives, without over-constraining or distorting sketch geometry. Using alignment techniques to train an existing constraint generation model with feedback from a constraint solver, we are able to fully-constrain 93\% of sketches compared to 34\% when using a naïve supervised fine-tuning (SFT) baseline and only 8.9\% without SFT. Our approach can be applied to any existing constraint generation model and sets the stage for further research bridging alignment strategies between the language and design domains. Additional results can be found at \url{https://autodeskailab.github.io/aligning-constraint-generation/}.
\end{abstract}
\vspace{-1em}
\section{Introduction}
\label{sec:introduction}

A central challenge in artificial intelligence (AI) is alignment: ensuring that AI systems produce outputs that adhere to human goals and expectations \cite{Russell2010, Christiano2017, Ouyang2022}. Although alignment of language models has been researched extensively \cite{Ouyang2022, rafailov2023direct, shao2024deepseekmath}, the application of alignment techniques to parametric design problems has yet to be studied. The use of AI in this discipline covers a broad range of areas, ranging from floor-plan layout~\cite{nauata2020house, shabani2023housediffusion}, to engineering design problems \cite{willis2020fusion, chen2020airfoil, regenwetter2022deep}, to 3D generation~\cite{xu2023hierarchical}. Design problems are often visual in nature and incorporate other functional requirements, making them unique when compared with language model alignment. In this paper, we establish the problem of \textit{design alignment} and demonstrate how this expansive problem can be made tractable by adapting techniques from the alignment literature into a new context.

In language models, alignment is achieved by incorporating feedback to generate coherent, contextually appropriate responses.  Similarly, with parametric CAD modeling, AI tools must be aligned with a designer’s intent by maintaining the underlying structural relationships to produce outputs that are both meaningful and functional. Otey et al.~\cite{Otey2018} define design intent as ``a CAD model’s anticipated behavior when altered,'' while Martin~\cite{Martin2023} characterizes it as ``relationships between objects, so that a change to one propagates automatically to others.'' This means that modifications of a design by an AI system should yield outcomes where the established design relationships remain intact. 
To that end, we define \textit{design alignment} as the application of generative modeling alignment techniques to produce outcomes that maintain design intent.

The realization of AI systems that observe and maintain design intent has broad implications for the manufacturing and construction industries. Almost every manufactured object or structure begins as a CAD model. At the core of parametric CAD modeling are 2D engineering sketches, which can be extruded or revolved to generate 3D models. Engineering sketches are composed of geometric primitives, such as points, lines, and circles, that are organized using constraints and dimensions~\cite{bouma1995}. These constraints\footnote{Throughout this paper, the term ``constraints'' is used in a broad sense to include both constraints (e.g., parallel) and dimensions (e.g., diameter).} define geometric rules, including equality, perpendicularity, and radial or linear dimensions, which collectively shape the final layout of the sketch. When applied correctly, they enable efficient modifications while preserving the original design intent.

\begin{figure}
    \begin{center}
        \includegraphics[width=\columnwidth]{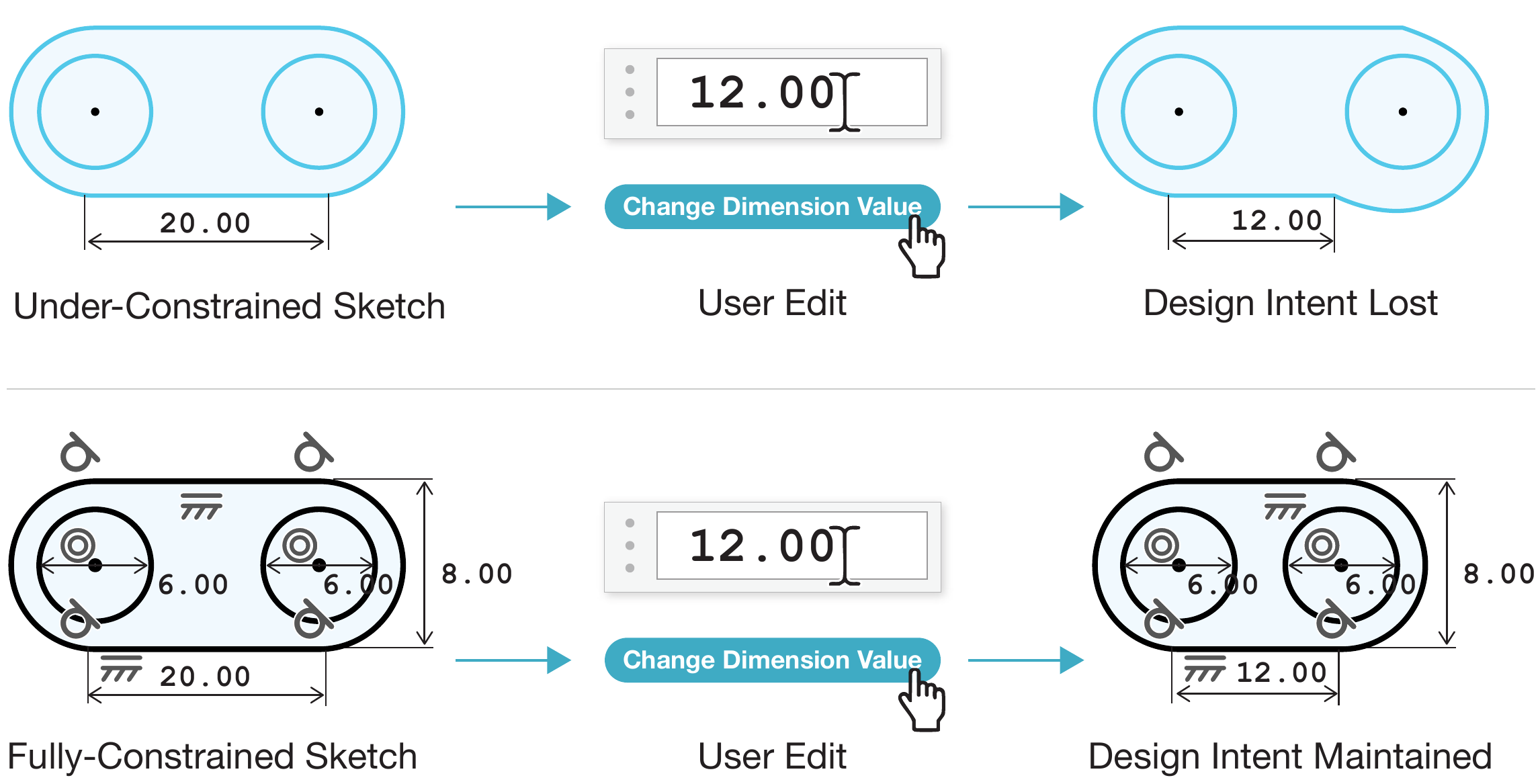}
        \caption{An illustration of design intent in CAD modeling. The bottom sketch maintains symmetry after modifying a dimension due to properly applied constraints, while the top sketch, lacking adequate constraints, becomes asymmetrical and distorted.}
        \label{fig:design_intent}
        \vspace{-2em}
    \end{center}
\end{figure}

Figure~\ref{fig:design_intent} illustrates the impact of constraint quality: a poorly constrained sketch loses symmetry when a dimension is changed, whereas a well-constrained sketch preserves its intended relationships. This underscores the important role of constraints in maintaining design intent, and the need for AI systems that can align with this intent encoded in designs.
We focus on the problem of sketch constraint generation~\cite{seff2020sketchgraphs} to demonstrate the adaption of alignment techniques to a design problem. Using an existing sketch constraint generation model Vitruvion~\cite{seff2021vitruvion}, we align the model with algorithms that learn from feedback (Direct Preference Optimization~\cite{rafailov2023direct}, Expert Iteration~\cite{anthony2017thinking, singh2024beyond}, RLOO~\cite{ahmadian2024rloo}, ReMax~\cite{li2024remax} and Group Relative Policy Optimization~\cite{shao2024deepseekmath}) using the sketch constraint solver in Autodesk Fusion \cite{fusion360sketch} as the learning signal. We optimize the models to remove all degrees of freedom in the sketches to become `fully-constrained'~\cite{Bernhard2011}, without causing sketches to be distorted, over-constrained or unsolvable. We further define these conditions in Section~\ref{sec:problem}.

To the best of our knowledge, this is the first instance of alignment methods being successfully applied to a parametric CAD design task; representing an important step forward for AI-assisted design tools. We present the following contributions:

\begin{itemize}
    \item We establish the problem of \textit{design alignment}, in the context of generative CAD models, as a critical component of AI-assisted CAD tools. We focus on the necessary first step of alignment for engineering sketches.
    \item We introduce a post-training strategy for a sketch constraint generation model using feedback from a sketch constraint solver. We define novel metrics and reward functions that directly optimize a base model for improved alignment.
    \item We conduct extensive experiments and demonstrate alignment techniques that fully-constrain 93\% of sketches compared to 34\% when using a naïve supervised fine-tuning (SFT) baseline and only 8.9\% without alignment. We posit that our approach is broadly applicable to other design tasks that require compilation of elements with rule-based algorithms.

\end{itemize}

\section{Related Work}
\label{sec:related-work}

\PAR{Engineering Sketches}
Engineering sketches form the 2D basis for 3D CAD models used to design mechanical parts for manufacturing. The availability of engineering sketch datasets~\cite{willis2020fusion, seff2020sketchgraphs, ganin2021computer} has enabled the development of generative models~\cite{willis2021engineering, ganin2021computer, seff2021vitruvion, para2021sketchgen} that can predict sketch geometry and/or the underlying constraints and dimensions that encode design intent. These Transformer-based~\cite{vaswani2017attention} approaches create geometry by autoregressively generating tokens representing points and curves, then add constraints by referencing this geometry using Pointer Networks~\cite{vinyals2015}. More recent approaches leverage image-based guidance~\cite{karadeniz2024davinci, wu2024cadvlm} or large language models (LLM)~\cite{jones2025solver} in the constraint prediction task. \citet{yang2022discovering} learn to group together recurring patterns of geometric and constraint entities within a sketch, effectively discovering latent design concepts. However, none of these approaches explicitly optimize for preserving design intent -- as a result, generated sketches may require additional manual refinement to capture the designer's intent.

Our work builds upon these foundations by explicitly incorporating design intent as a post-training process. Instead of merely modeling the ground truth data, our method learns from constraint solver feedback, ensuring that generated sketches are geometrically plausible and structurally well-constrained. By doing so, we enable data-driven generation that aligns with design intent.

\PAR{Design Alignment}
Beyond the language domain, alignment techniques have been used to improve and align image generation.
\citet{lee2023aligning} propose fine-tuning diffusion-based text-to-image models using human feedback, significantly improving alignment between textual prompts and generated visuals. Similarly, ImageReward~\cite{xu2023imagereward} uses a learned reward model trained on human preference data, which guides the diffusion model fine-tuning toward images preferred by human evaluators. Extending this idea, \citet{black2023training} reframes image generation as a sequential RL task, introducing Denoising Diffusion Policy Optimization (DDPO) to optimize complex user-defined objectives directly for alignment without explicit human annotation. 
Few works have applied alignment techniques in the design domain. GearFormer~\cite{etesam2024gearformer} used differentiable sampling to enforce preferences in the solutions for mechanical configuration design problems, however, this approach does not work with rewards that require blackbox solvers in the loop. In concurrent work, e-SimFT~\cite{cheong2025simft} used preference data obtained from physics simulations to enhance the exploration of the Pareto front in a multi-preference setting, improving the solutions generated with GearFormer.

\PAR{Fine-tuning LLMs with RL}
Reinforcement Learning from Human Feedback (RLHF) has emerged as a cornerstone approach for aligning large language models (LLMs) with human preferences. In RLHF~\cite{Christiano2017,Ouyang2022, bai2022constitutional, schulman2017proximal}, a learned reward model is usually trained to capture human preferences and is provides the learning signal that the policy is trained on. Other methods such as Direct Preference Optimization (DPO)~\cite{rafailov2023direct} and Reinforced Self-Training (ReST)~\cite{singh2024beyond} use a simpler approach of optimizing the model directly from model generated data without the use of a learned reward model. While DPO learns from ranked pairs of generated model outputs from human annotators, ReST uses rejection sampling to remove incorrect generations and trains the model standard cross-entropy loss on the correct samples. In this paper, we refer to the approach of using fine-tuning on high return responses as Expert Iteration (ExIt). We broadly refer to algorithms that learn from ranked/filtered model outputs (such as DPO and Expert Iteration) as Preference Optimization (PO).

More recently, a large body of work has focused on the task of teaching LLMs to solve reasoning tasks with reinforcement learning~\cite{havrilla2024teaching, lambert2024t, singh2024beyond, gehring2024rlef}, such as math and coding, which can be checked with rule-based systems. Specifically we are inspired by approaches which forego the use of a learned reward model and directly learn from verifiable rewards. We build off of several approaches that have shown success when applied to reasoning LLMs -- these include Group Relative Policy Optimization (GRPO)~\cite{shao2024deepseekmath}, ReMax~\cite{li2024remax}, and Reinforce Leave-One-Out (RLOO)~\cite{ahmadian2024rloo}. Both GRPO and RLOO estimate the baseline via the average reward of multiple sampled outputs instead of learned value model but differ in how they apply the KL divergence penalty, advantage normalization, and PPO-style reward clipping. ReMax~\cite{li2024remax} also obviates the need for a learned value model but instead estimates the baseline from the argmax result (greedy sampling).

In this paper, we adapt the aforementioned post-training methods---DPO, Expert Iteration, RLOO, GRPO, ReMax, for use in aligning a constraint generation model using feedback from a constraint solver. In \Cref{sec:method} we provide additional details on the ranking/filtering criteria for the Preference Optimization (PO) algorithms and the reward design for the RL algorithms (GRPO, RLOO, ReMax).

\section{Problem}
\label{sec:problem}

Sketch constraining is a fundamental component of parametric CAD modeling, where geometric relationships define the structure and behavior of the sketch. Applying constraints ensures stability and editability, allowing for parametric modifications that align with the design intent. Automating constraint generation requires producing a valid and efficient set of constraints that fully define a given sketch while avoiding unnecessary redundancy or conflicts.

The sketch constraining problem can be formulated as a sequence modeling task similar to natural language generation, where constraints are predicted autoregressively. Given the sketch geometry as input, the model generates a sequence of constraints in the order they will be applied. Tokens in the sequence represent either a constraint (e.g., coincident, parallel, perpendicular), a dimension (e.g., horizontal, vertical, radial), or a pointer to one of the input geometric entities~\cite{vinyals2015}. 

In parametric CAD, sketch geometry is modified using a constraint solver. The updated sketch respects any present constraints while moving the geometry to reflect changes to the dimension parameters. Unlike natural language, which is inherently sequential and follows flexible grammar rules, sketch constraints must adhere to strict geometric principles to ensure structural validity. A set of constraints may be incorrect for a variety of reasons: they can reference the wrong primitives for the constraint type, be redundant, specify inconsistent geometric relationships, or cause unexpected geometric distortions.

\begin{figure}[t]
    \centering
    \includegraphics[width=\linewidth]{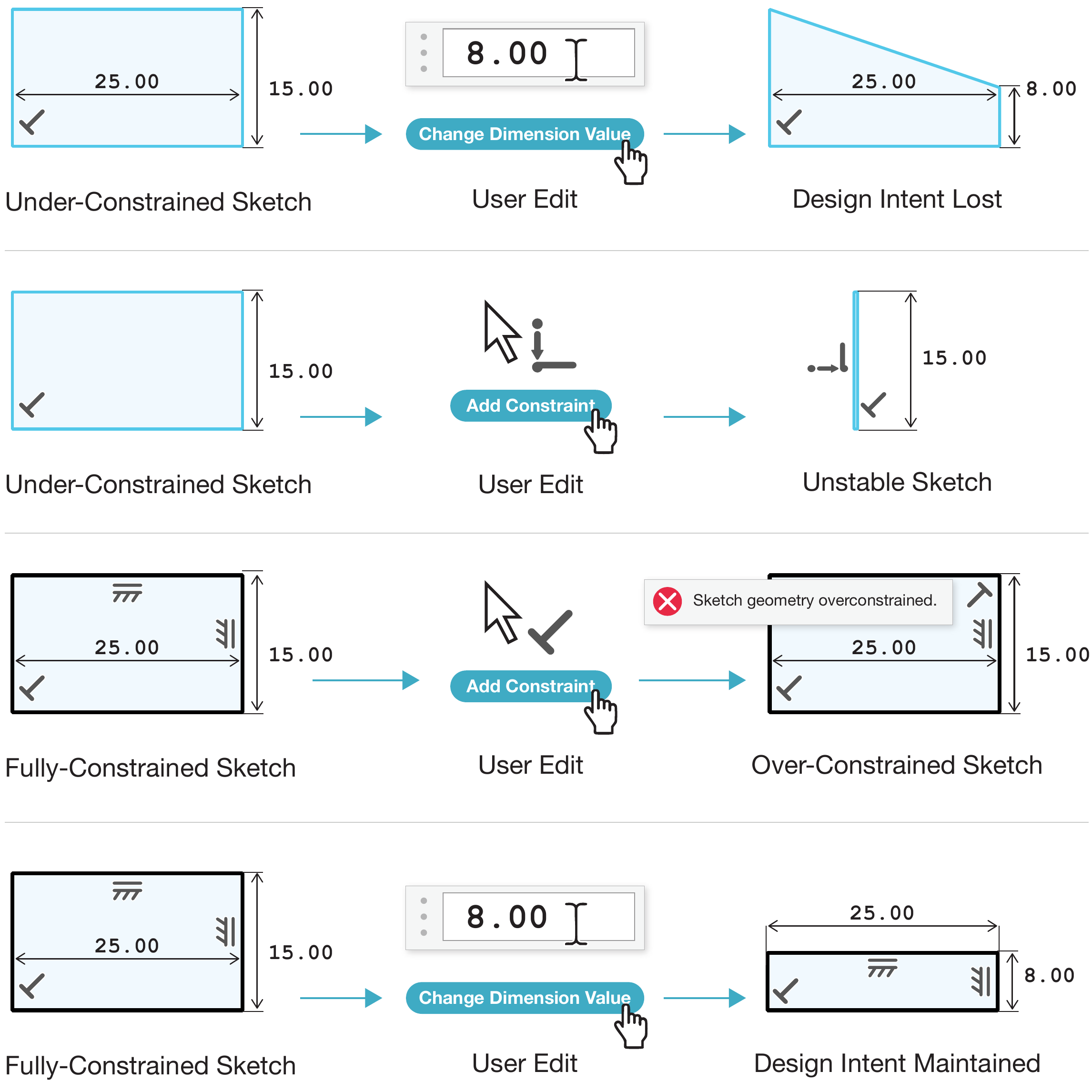}
    \caption{Comparing different outcomes when a designer updates a sketch parameter on the constrained sketch. \textbf{First row:} an under-constrained sketch only preserves a subset of the geometric relationships. \textbf{Second row:} the sketch is unstable, adding a coincident constraint flattens the geometry of the sketch. \textbf{Third row:} the sketch is over-constrained, causing the sketch to be uneditable. \textbf{Fourth row:} all geometric relationships are maintained after the parameter is updated. }
    \label{fig:fcoc_visual}
    \vspace{-1em}
    
\end{figure}

\begin{figure*}[t]
    \centering
    \includegraphics[width=\textwidth]{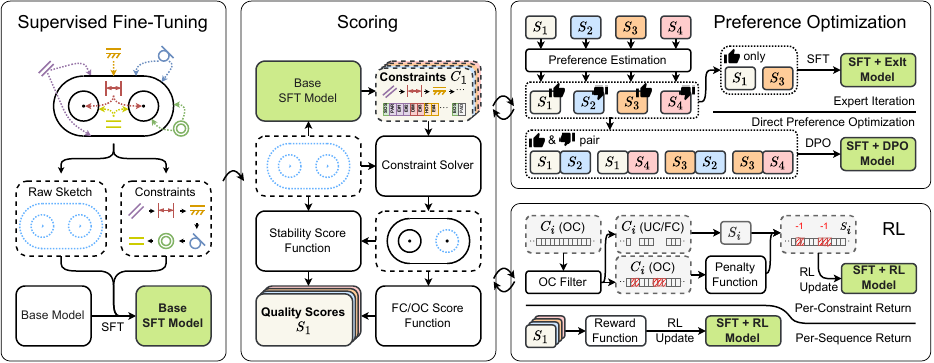}
    \caption{Illustration of the proposed alignment workflow for constraint generation models. \textbf{Left:} A base constraint-generation model is first fine-tuned using supervised fine-tuning (SFT). \textbf{Middle:} Generated constraint sequences $C_i$ are evaluated using a constraint solver, which provides feedback on sketch stability and fully- and over-constrained statuses, forming the quality score vector $S_i$. \textbf{Right:} Two groups of alignment methods leveraging solver feedback: preference-based optimization (PO) and reinforcement learning (RL). PO uses $S_i$ to construct training data, iteratively improving constraint prediction quality. RL methods assign per-constraint and per-sequence rewards to $C_i$ based on solver feedback and $S_i$ to incentivize the generation of constraints, making sketches stable and fully-constrained.}
    \label{fig:system}
    \vspace{-1em}
\end{figure*}

We formally define five conditions describing the state of a sketch after applying constraints. These conditions are not mutually exclusive. A sketch may satisfy one or more conditions depending on the applied constraints. Figure~\ref{fig:fcoc_visual} provides illustrative examples of each condition.

\PAR{Under-constrained (UC)} A sketch containing primitives retains some unconstrained degrees of freedom, resulting in incomplete specification of their positions or dimensions.

\PAR{Fully-constrained (FC)} A sketch in which all primitives have their degrees of freedom completely determined, removing positional or dimensional ambiguities.

\PAR{Over-constrained (OC)} A sketch primitives have more constraints applied than degrees of freedom, potentially leading to conflicts. Note some over-constrained sketches remain solvable if the constraints are consistent and do not conflict with each other~\cite{bouma1995}.

\PAR{Not solvable} A sketch that cannot achieve a valid solution due to contradictory or redundant constraints, leading to an impossible or conflicting geometry.

\PAR{Stability} We discretize the sketch plane into a grid and classify a sketch as \textbf{unstable} if the positions of primitives after constraint solving shift into different cells. The number of cells (bins) on each axis determines the sensitivity of this measurement. 

Our objective is to align the model toward generating constraint sets that yield fully-constrained sketches while minimizing cases of under-constrained, over-constrained, not solvable, or instability. This is a necessary prerequisite toward the ultimate goal of generating constraints that preserve the original sketch design intent when the designer varies the geometric parameters. A formal definition of FC, UC, and OC is described in ~\cite{hoffmann2005brief,hoffmann1997symbolic} and in the appendix.

\section{Method}
\label{sec:method}

In this section, we outline the post-training techniques used for aligning a constraint generation model with feedback from a constraint solver. The approaches are grouped into three categories: supervised learning methods, preference-based optimization methods, and RL methods. \Cref{fig:system} illustrates the high-level workflow of this work.

\subsection{Constraint Solver}

We evaluate generated constraints using the commercially available constraint solver in Autodesk Fusion. This solver is treated as a black box, from which we retrieve sketch conditions as defined in~\Cref{sec:problem}. A general algorithmic approach commonly used in constraint solvers is described in~\cite{bouma1995geometric}. The solver outputs the fully-constrained status for each entity if the sketch is solvable. Otherwise it classifies the sketch as over-constrained or unsolvable. The solver adjusts geometry to resolve all constraints which we compare with the original sketch to determine if the constraints caused geometry distortion (instability). On average, it takes 0.1–0.2 seconds to per solve, although complex cases may take tens of seconds. Any sketch taking longer than two seconds to solve is automatically deemed unsolvable.

\subsection{Supervised Learning}
We pre-train Vitruvion~\cite{seff2021vitruvion} as our base model using the same procedure described in their paper with a next-token prediction objective. Given an input sequence of geometric primitives, the model is trained to predict the next correct constraint or dimension based on the ground truth data. Further details about our implementation of the Vitruvion architecture and training procedure are provided in the appendix.

Since the majority of the ground truth data contains under-constrained or over-constrained sketches, we additionally perform supervised fine-tuning (SFT). During SFT, the training data is limited to sketches verified by the constraint solver as solvable, fully-constrained, stable, and free of over-constrained conditions, ensuring the model explicitly learns from ideal examples.

\subsection{Preference-Based Optimization (PO)}

\PAR{Expert Iteration (ExIt)} alternates between expert improvement and policy distillation. Following \cite{singh2024beyond, havrilla2024teaching}, we use temperature sampling combined with rejection sampling to generate high-quality candidate constraint sequences. During the exploration step we initialize the policy model \(\pi_{\theta_{t=0}}\) from the SFT model and sample $K = 8$ candidate constraint sequences $\tau$ at temperature $T = 1.0$ for each sketch query $q$ in the initial training set ${\mathcal{D}}$. A new training dataset $\mathcal{D}^*_{i \in {N}}$ is constructed by discarding sequences that are under-constrained, over-constrained, or unsolvable solutions. This process is repeated \(N = 2\) times over the dataset, and the policy is trained using cross-entropy loss:  

\begin{equation}
    \mathbb{E}_{(q, \tau) \sim \mathcal{D}^*_{i \in {N}}} \left[ \log \pi_{\theta_t}(\tau | q) \right]
\end{equation}
where $\pi_{\theta_t}$ is updated after the distillation phase of every training round.

\PAR{Direct Preference Optimization (DPO)} learns from pairwise preference data, approximating an implicit reward via a reparameterized Bradley-Terry model~\cite{rafailov2023direct}. Similar to ExIt, we construct the training dataset by sampling \(K = 8\) constraint sequence completions \(\tau\) for each sketch query \(q\) from the policy model \(\pi_{\theta_t}\) at temperature $T=1.0$. Pairs \((\tau_w, \tau_l)\) are ranked based on the differences in the percentage of fully-constrained curves between $\tau_w$ (preferred sequence), and $\tau_l$ (non-preferred sequence). 
The optimization objective is:
\begin{equation}\label{eq:optimum_model}
    \mathbb{E}_{(q, \tau_w, \tau_l) \sim \mathcal{D}} \left[
    \log \sigma \left(
    \beta \log \frac{\pi_{\theta_t}(\tau_w|q)}{\pi_{\theta_r}(\tau_w|q)} 
    - \beta \log \frac{\pi_{\theta_t}(\tau_l|q)}{\pi_{\theta_r}(\tau_l|q)}
    \right) \right]
\end{equation}
where \(\beta\) is a hyperparameter, and $\sigma$ is the logistic function. This formulation ensures the learned policy \(\pi_\theta\) aligns with the ranking preference of fully-constrained sketches while maintaining proximity to the reference policy \(\pi_{\theta_r}\).

We initialize \(\pi_{\theta_{t=0}}\) from the SFT model and repeat the entire process $N = 2$ times, updating \(\pi_{\theta_t}\) with the new policy after every training iteration. Additional hyperparameters and ranking details can be found in the appendix.


\subsection{Reinforcement Learning (RL)}

\subsubsection{Reward design}
\label{sec:reward_design}
Unlike natural language tasks, where human preferences are ambiguous and ill-defined, the stability and solvability of sketch constraints can be verified, making it compatible with RL methods that directly optimize for mechanically defined rewards without a learned preference model. 
We define the rewards used for RL as follows:

\begin{itemize}
    \item Rewards for valid constraint sequence $\tau$:
    \begin{description}
    \item [$r_{\text{curves}}(\tau)$:] \% of fully-constrained curves over all curves,
    \item [$r_{\text{points}}(\tau)$:] \% of fully-constrained points over all points,
    \item [$r_\text{unstable}$:] penalty for unstable sketches,
    \end{description}
    \item Rewards for invalid constraint sequence:
    \begin{description}
    \item [$r_\text{NS}$:] penalty for not solvable sketches,
    \item [$r_\text{OC}$:] penalty for over-constrained sketches,
    \item [$r_\text{F}$:] penalty for sketches resulting in other failures.
    \end{description}
\end{itemize}
The overall sequence-wise reward $R(\tau)$ is the sum of $r_{\text{curves}}(\tau)$, $ r_{\text{points}}(\tau)$, and conditionally $r_\text{unstable}$ for valid sketches, and either $r_\text{NS}$, $r_\text{OC}$ or $r_\text{F}$ for invalid sketches according to the failure mode.

We additionally define a constraint-wise penalty to provide granular feedback on cases where the constraint sequence causes sketches to be over-constrained or fully-constrained. A constraint solver iteratively attempts to add each generated constraint one-by-one, dropping any problematic constraints that caused the sketch to be over-constrained or not-solvable. In training, we add a constant of -1 loss penalty directly to the per-token log likelihood loss for the problematic constraints.

\subsubsection{RL fine-tuning formulation}
\label{sec:rl_problem}
We formulate constraint generation fine-tuning as follows; given a dataset of sketch queries $D = \{q_i\}_{i=1}^N$ and reward function $R(\tau)$ using the constraint solver and reward design in \Cref{sec:reward_design}, learn a policy $\pi_\theta(\tau|q)$ that generates a sequence of constraints $\tau$ for sketch $q$, such that it maximizes the expected rewards $\mathbb{E}_{q_i\sim D, \tau_i \sim \pi_\theta(\cdot|q_i)}[R(\tau_i)]$.

\subsubsection{Policy gradient methods}
For RLHF which uses a pre-trained policy, not all the complexity of RL algorithms is necessary. This allows the algorithms to be simplified and the number of learnable components to be reduced, contributing to performance improvement. We considered three policy gradient algorithms: ReMax~\cite{li2024remax}, RLOO~\cite{ahmadian2024rloo}, and GRPO~\cite{shao2024deepseekmath}. Unlike PPO~\cite{schulman2017proximal}, which treats each token generation as an action, these algorithms treat generation of a sequence as a single action and adopt a REINFORCE with baselines approach~\cite{lex2001reinforcebaseline}. We apply these to optimize the constraint generation policy. 

\PAR{ReMax}\cite{li2024remax} uses the rewards corresponding to sequences generated by a greedy (argmax) policy as a baseline to normalize the rewards of sequences sampled from the policy. 

Considering sequence $\tau$ sampled from policy $\pi_{\theta}(\tau | q)$, sequence $\tau^* = \text{argmax}_{\tau} \pi_{\theta}(\tau | q)$ greedily sampled by taking an argmax of the policy, and corresponding rewards $r$ and $r^*$, the policy gradient objective for ReMax is:

\begin{equation}
    \mathop{\mathbb{E}}_{\tau \sim \pi} \left[ (r -  r^*) \nabla \log \pi_\theta(\tau | q) \right].\\
\end{equation}

\PAR{REINFORCE-Leave-One-Out (RLOO)}\cite{ahmadian2024rloo} samples $G$ number of constraint sequences for every sketch query. The baseline for each sample is evaluated as the mean of the rewards for all other samples in the group.  

For $G$ number of sequences $\{\tau_{g}\}_{g=1}^G$ sampled from policy $\pi_{\theta}(\tau | q)$ for a given sketch query $q$, the policy gradient objective of RLOO is:

{\small
\begin{equation}
    \mathop{\mathbb{E}}_{\{\tau\} \sim \pi} \left[ \frac{1}{G}\sum_{g=1}^G  \left[\left(r_g - \text{mean}(\{r_i\}_{i\neq g}^G) \right) \nabla \log \pi_\theta(\tau_g | q_g) \right] \right].\\
\end{equation}}

\PAR{Group Relative Policy Optimization (GRPO)}\cite{shao2024deepseekmath} uses group-based baseline estimation, like RLOO, but the mean is taken over all reward samples in the group. It uses a clipped policy optimization objective similarly to PPO~\cite{schulman2017proximal}, as well as a low-variance KL regularization term.

For $G$ number of sequences $\{\tau_{g}\}_{g=1}^G$ sampled from the reference policy $\pi_{\theta_\text{r}}(\tau | q)$ for a given sketch query $q$, letting $\rho_g = \frac{\nabla \pi_\theta(\tau_g|q)}{\pi_{\theta_\text{r}}(\tau_g|q)}$, the optimization objective of GRPO is:
{\small
\begin{align}
\begin{split}
    &\mathop{\mathbb{E}}_{\{\tau_g\} \sim \pi} \Bigl[ \min \left( \rho_g A_g,\text{clip}\left( \rho_g, 1-\epsilon, 1+\epsilon \right) A_g \right) - \beta \mathbb{D}_\text{KL}(\pi_\theta || \pi_{\theta_\text{r}}) \Bigr],\\
    &\mathbb{D}_\text{KL}(\pi_\theta || \pi_{\theta_\text{r}}) = \frac{1}{\rho_g} + \log\rho_g - 1, \quad A_g=\frac{r_g - \text{mean}(\{r_g\}_{g=1}^G)}{\text{std}(\{r_g\}_{g=1}^G)},
\end{split}
\end{align}}
where $\epsilon$ and $\beta$ are hyper-parameters for the clipped policy optimization and KL regularization terms, respectively. 

\begin{table*}[ht]
    \centering
    \small
    \caption{Sketch constraint generation results for Fully Constrained (FC), Under Constrained (UC), Over Constrained (OC), not solvable, and stability for the base model, SFT model, and aligned models. Results are computed over 8 samples per sketch with a temperature of 1.0. Numbers following $\pm$ indicate the standard deviation.}
    \begin{tabular}{lccccc}
        \toprule
        Model & \% FC $\uparrow$ & \% UC $\downarrow$ & \% OC $\downarrow$ & \% Not solvable $\downarrow$ & \% Stable (bins=4) $\uparrow$ \\
        \midrule
        Vitruvion (base)     &  8.87 {\scriptsize$\pm$0.09}  & 71.38 {\scriptsize$\pm$0.20}  & 16.83 {\scriptsize$\pm$0.17}  & 3.05 {\scriptsize$\pm$0.03}  & 92.15 {\scriptsize$\pm$0.06}  \\
        SFT                  & 34.24 {\scriptsize$\pm$0.09}  & 46.61 {\scriptsize$\pm$0.15}  & 15.30 {\scriptsize$\pm$0.08}  & 3.85 {\scriptsize$\pm$0.03}  & \bf{92.48} {\scriptsize$\pm$0.05}  \\
        Iterative DPO        & 64.91 {\scriptsize$\pm$0.11}  & 14.97 {\scriptsize$\pm$0.13}  & 12.47 {\scriptsize$\pm$0.09}  & 7.64 {\scriptsize$\pm$0.07}  & 87.63 {\scriptsize$\pm$0.09}  \\
        Expert Iteration     & 71.70 {\scriptsize$\pm$0.13}  & 13.38 {\scriptsize$\pm$0.13}  & 7.25 {\scriptsize$\pm$0.06}   & 7.67 {\scriptsize$\pm$0.07}  & 85.50 {\scriptsize$\pm$0.10}  \\
        ReMax                & 79.84 {\scriptsize$\pm$0.09} & 15.86 {\scriptsize$\pm$0.07} & \bf{1.49 {\scriptsize$\pm$0.01}}  & 2.82 {\scriptsize$\pm$0.02}   & 75.77 {\scriptsize$\pm$0.05}          \\
        RLOO                 & \bf{93.05 {\scriptsize$\pm$0.03}}  & \bf{3.55 {\scriptsize$\pm$0.02}}   & 2.15 {\scriptsize$\pm$0.01}   & \bf{1.25 {\scriptsize$\pm$0.01}}   & 89.16 {\scriptsize$\pm$0.02}  \\
        GRPO                 & 91.59 {\scriptsize$\pm$0.03}  & 4.18 {\scriptsize$\pm$0.03}   & 1.94 {\scriptsize$\pm$0.01}   & 2.28 {\scriptsize$\pm$0.02}   & 88.28 {\scriptsize$\pm$0.03}  \\
        \bottomrule
    \end{tabular}
    \label{tab:main_results}
\end{table*}

For ReMax and RLOO, we also added a small KL penalty term to the rewards to discourage divergence from the reference policy. GRPO applies group-normalization on the advantage, which we also applied in RLOO. For ReMax, we batch-normalized the advantage. For all algorithms, $\pi_\theta$ is initialized from the SFT model. Further implementation details of the algorithms can be found in the appendix.

\section{Experiments}
\label{sec:experiments}

\subsection{Dataset}

We train our models on SketchGraphs~\cite{seff2020sketchgraphs}, a large-scale dataset of CAD sketches created in Onshape. SketchGraphs captures real-world parametric modeling workflows, providing geometry and constraint construction operations from actual design steps. However, the dataset was not originally designed for direct constraint inference; only 8.27\% of its sketches are fully-constrained, making it imperfect for training the constraint generation model.

For computational feasibility, we deduplicate and filter sketches, retaining only those with at most 16 geometric primitives and 64 constraints, yielding a dataset of 2.8 million unique sketches. Certain constraint types, such as symmetry, are excluded to simplify the learning task, ensuring the focus remains on constraints most relevant to engineering design. We also convert Onshape sketches into the Fusion sketch format to utilize the solver in Fusion. Additional details are provided in the appendix. 

Another challenge lies in how SketchGraphs positions primitives. Rather than being placed in valid, constraint-satisfying layouts, primitives often have arbitrary coordinates that do not reflect a solved state. We therefore preprocess the dataset using Fusion to resolve each sketch’s primitives according to its constraints, ensuring that geometry and constraints match before training.

\subsection{Quantitative Results}

In~\Cref{tab:main_results}, we list results comparing the performance of each alignment method with respect to the five sketch conditions described in~\Cref{sec:problem}. The base model is able to fully-constrain sketches only 8.87\% of the time, consistent with the dataset distribution where only 8.27\% of sketches are fully constrained. We find that RLOO and GRPO perform similarly, giving the best performance at fully-constraining sketches 93.05\% and 91.59\% of the time, respectively. They have the lowest indicents of over-constrained or unsolvable results and maintain stability rates that are on par with other methods.

\begin{table}[b]
    \centering
    \small
    \caption{Sketch constraint generation results for Pass@1 and Pass@8 across the post-training algorithms. We define a successful result as fully-constrained, not over-constrained, solvable, and stable at 4 bins. Results are generated with temperature of 1.0.}
    \begin{tabular}{lcc}
        \toprule
        Model                & Pass@1 & Pass@8 \\
        \midrule
        Vitruvion (base)     & 8.53  & 20.47    \\
        SFT                  & 33.32 & 42.62    \\
        Iterative DPO        & 59.38 & 68.32   \\
        Expert Iteration     & 64.09  & 72.56   \\
        ReMax                &  62.74 &  65.89      \\
        RLOO                 & \bf{83.57} & \bf{84.96}    \\
        GRPO                 & 81.49 & 83.42  \\
        \bottomrule
    \end{tabular}
    \label{tab:pass_n_results}
\end{table}

\begin{figure*}[]
    \centering
    \includegraphics[width=\textwidth]{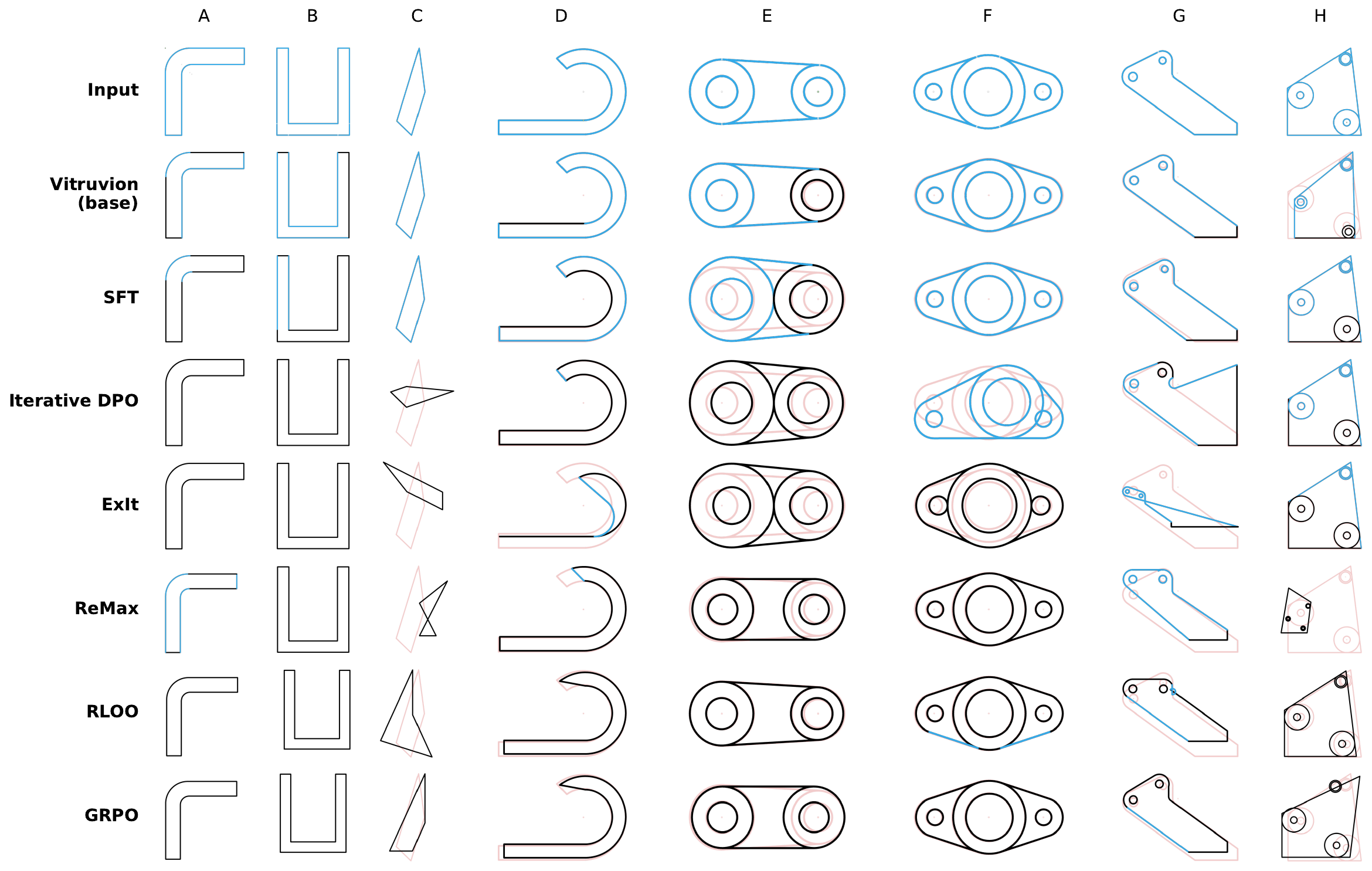}
    \caption{Visual comparison of solved sketches across the aforementioned baseline and post-training methods. Curves are colored black when fully-constrained and blue when not. Overlaid red lines are the original input sketch for reference.}
    \label{fig:qual_results}
    \vspace{-1em}
\end{figure*}

Iterative DPO and ExIt significantly improve upon the base and SFT models but still fall short of the performance achieved by policy gradient-based RL methods. We attribute this gap to the online nature of policy gradient-based RL, which continuously refines the policy through feedback while actively exploring a broader range of solutions. In contrast, Iterative DPO and ExIt are offline methods and rely on predefined ranking and filtering signals to generate training data, which limits their ability to explore the solution space. The superior performance of online RL underscores its advantage in directly optimizing the shaped rewards from the constraint solver.

\Cref{tab:pass_n_results} demonstrates the evaluation result of our methods in a few-shot inference setting, where the model has $K$ attempts to generate a fully-constrained and stable output. This scenario reflects real-world use cases where the goal is to maximize the likelihood of producing an acceptable solution within a fixed inference budget of $K$ samples (Pass@$K$). We find that while the RL-based methods still have the overall highest performance, increasing the number of samples $K$ has a comparatively small impact on performance compared to the other methods. Additional results and analysis on the impact of sampling parameters such as temperature and top-$p$ are available in the appendix.

\subsection{Qualitative Results}

We randomly select sketches from the test set and show the results generated by different alignment methods in ~\Cref{fig:qual_results}. Curves are colored black when constrained and blue when not. Methods leveraging alignment techniques demonstrate a promising trend towards generating fully-constrained sketches, whereas the base Vitruvion and SFT models typically leave sketches under-constrained. However, across different alignment algorithms, we observe substantial variance in the degree of geometric distortion introduced by the aligned models.

Specifically, columns A and B depict simple sketches mainly composed of horizontal and vertical lines, for which all solver-feedback methods consistently yield fully-constrained, stable results. In contrast, Column C presents a challenging sketch due to the absence of appropriate constraints for oblique lines. Achieving stability in this scenario typically requires many dimensions but few constraints; however, models are optimized to generate more constraints to align with parametric CAD design principles.

Sketches in columns D through H include arcs. Column E is particularly notable as all solver-feedback methods achieve a fully-constrainted condition, yet only RLOO produces visually stable results. Further examination reveals that distortions induced by other models were still classified as stable due to our bin size ($bins=4$). Column G presents unique challenges with an isolated point, causing confusion in the model and demanding extensive use of non-horizontal/vertical constraints to make the sketch fully-constrained and stable.

Our results indicate that models more easily fully-constrain sketches consisting primarily of horizontal and vertical lines without distortion. In contrast, sketches involving oblique lines increase the challenge of maintaining stability, and arcs further complicate achieving fully-constrained status. Among the tested methods, RLOO and GRPO exhibit the strongest overall performance in achieving fully-constrained status and geometric stability. 

Upon closer inspection of the generated constraints for the RL-based methods, we find that these models learns to ``reward hack" by adding far more dimensions than constraints. This allows the model to avoid the instability penalty because dimensions cannot cause geometry distortion. In order to counteract this behavior, we added an additional reward penalty on the ratio of constraints versus dimensions which fixes the over-dimensioning problem at the cost of reducing stability and fully-constrained status. Additional details are provided in the appendix.

To further evaluate the efficacy of each method, we conduct a forced-choice perceptual study with professional CAD designers. We show them paired images of the same sketch with different constraints applied from each method, to understand which set of constraints they prefer. We find that the preference-based optimization methods (ExIT, DPO) and RLOO with the over-dimensioning penalty consistently outperform the SFT and vanilla RLOO results. Results of this study are provided in the appendix.
\section{Limitations}
\label{sec:limitations}

Our experiments focus on sketches of moderate complexity, with a maximum of 16 geometric primitives and 64 constraints. Evaluating performance on more complex sketches requires further exploration. Additionally, our alignment approach only uses feedback signals from a constraint solver to represent design intent. Incorporating subjective human preferences and explicit design intent into alignment objectives remains a promising area for future work.

\section{Conclusion}
\label{sec:conclusion}

We demonstrated the adaption of alignment strategies from language modeling to preserve design intent in parametric CAD sketches. By using feedback from a constraint solver  as a learning signal, we show the feasibility and value of alignment in parametric CAD tasks. In doing so, we pave  the way for future AI-assisted design tools that incorporate design alignment.

{
    \small
    \bibliographystyle{ieeenat_fullname}
    \bibliography{main}
}


\appendix
\renewcommand{\thefigure}{\Alph{section}.\arabic{figure}}
\renewcommand{\thetable}{\Alph{section}.\arabic{table}}

\cleardoublepage

\section*{Appendix}

\section{Parametric CAD Sketches}

We discuss additional details related to our dataset of parametric CAD sketches and constraint solver.

\setcounter{figure}{0}
\setcounter{table}{0}

\subsection{Background}
\label{appendix_background}
Parametric CAD fundamentally relies on sketches as the basis for generating complex 3D geometries. Sketches are formed from geometric primitives such as points, lines, arcs, and circles. By imposing constraints (e.g., tangency, perpendicularity, parallelism) and dimensions (e.g., linear, angular, radial), these primitives become systematically interlinked, preserving design intent through iterative modifications. A dedicated constraint solver manages this network of relationships, using numerical methods to maintain consistency and automatically adjust dependent elements when any single parameter changes.

In Figure~\ref{fig:constraint_refs}, tangent constraints (blue) reference a line and an arc, horizontal constraints (orange) reference two lines, and linear dimensions (red) reference two points. Such definitions encode both geometric relationships and key measurements, allowing the solver to propagate updates throughout the model. This approach reduces the need for manual rework by ensuring that changing one dimension, such as the distance between two points or the radius of an arc, will automatically update the entire sketch. This allows designers to iterate rapidly while maintaining the design intent embedded in the sketch.

\setcounter{figure}{0}
\setcounter{table}{0}

\subsection{Constraint State Definitions}
The formal characterization of sketch constraint states follow Hoffman~\cite{hoffmann2005brief, hoffmann1997symbolic}.  
Let a sketch be parameterized by a set of geometric parameters \(P = \{p_1, \dots, p_n\}\) and defined by a set of constraint equations \(C = \{c_1, \dots, c_m\}\).  
The Jacobian matrix \(J_C = \frac{\partial C}{\partial P}\) captures the local dependency between parameters and constraints.  
A sketch is \textbf{fully-constrained (FC)} if \(\text{rank}(J_C) = n - r\), where \(r\) represents the residual rigid-body degrees of freedom.
It is \textbf{under-constrained (UC)} if \(\text{rank}(J_C) < n - r\), implying that at least one geometric parameter retains an unconstrained degree of freedom.  
It is \textbf{over-constrained (OC)} if \(\text{rank}(J_C) > n - r\), indicating redundant or inconsistent constraints, which may still yield a solvable configuration if the constraints are algebraically consistent.

\subsection{Unstable Sketch Definition}
\begin{figure}
    \begin{center}
        \includegraphics[width=\columnwidth]{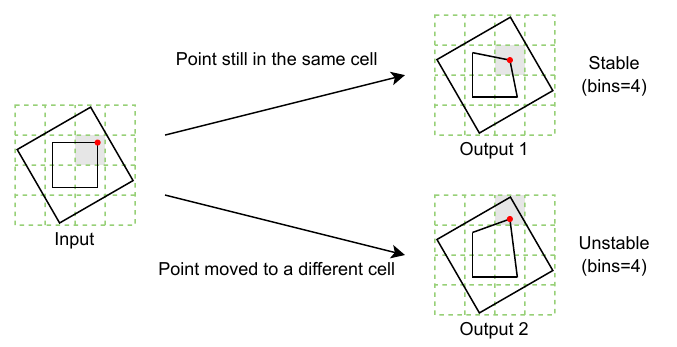}
        \caption{Visualization of stable versus unstable sketches using a $4 \times 4$ grid. Sketches with all points remaining in the same cell are considered stable (top), while those that move to a different cell are marked unstable (bottom).}
        \label{fig:unstable_example}
    \end{center}
\end{figure}

To evaluate whether a sketch remains geometrically stable after constraint application, we introduce a metric that detects significant shifts in the sketch geometry. Specifically, we divide the sketch canvas into an $n\times n$ grid of spatial bins as shown in~\Cref{fig:unstable_example}. A sketch is deemed \emph{unstable} if any of its geometric entities move from their original bin to a different bin after constraint solving. This condition implies a meaningful deformation rather than minor numeric jitter. Such instability may indicate poorly conditioned constraint sets, where the solver resolves constraints by distorting the geometry. We apply this rule to all generated outputs and classify each sketch as either stable or unstable.

\subsection{Fusion Sketch Representation}
\label{appendix_fusion}
\begin{figure}
    \begin{center}
        \includegraphics[width=\columnwidth]{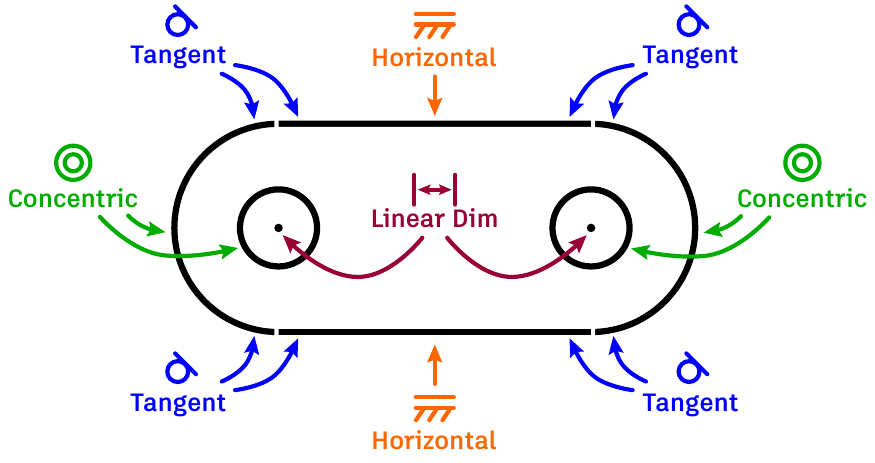}
        \caption{An example sketch illustrating how constraints and dimensions reference geometric primitives such as points, lines, arcs, and circles. A constraint solver enforces these relationships, ensuring that a change in one parameter propagates consistently throughout the sketch.}
        \label{fig:constraint_refs}
    \end{center}
\end{figure}

The Fusion 360 Gallery sketch format~\cite{willis2020fusion} organizes sketch elements into a hierarchical, structured representation, wherein a sketch is defined by a set of parametric geometric primitives and a set of explicit constraints between those primitives. Each geometric primitive (line, arc, circle, point, etc.) is described by its intrinsic parameters (e.g., endpoint coordinates for a line, center and radius for a circle). Alongside the primitives, the sketch includes constraints (e.g., coincident points, perpendicular or parallel lines) that impose geometric relationships to be satisfied simultaneously. These constraints serve to preserve design intent: for instance, a coincidence constraint can lock the endpoint of a line onto a circle’s circumference, or an equal-length constraint can enforce that two segments remain the same length.

Structuring the sketch with primitives and constraints yields a rich, relational format rather than a flat drawing. The representation can be viewed as a bipartite graph, where primitive nodes carry geometric parameters and constraint edges specify relationships linking one or more primitives.

\subsection{Sketch Tokenization}
\label{appendix_tokenization}
Our tokenization of sketches defines a diverse vocabulary of token types to represent the heterogeneous elements of a sketch. There are distinct token categories for primitive types, constraint and dimension types, and special markers (e.g., \texttt{<SOS>}, \texttt{<EOS>}, \texttt{<PAD>}). In our approach, constraint tokens, dimension tokens, and primitive reference tokens are the primary outputs of the model. These tokens are strictly categorical, reflecting the discrete nature of constraint types and their relationship to previously defined primitives. For example, a perpendicular constraint might be tokenized as (\texttt{<PER>}, \texttt{<REF\_A>}, \texttt{<REF\_B>}), where \texttt{<REF\_A>} and \texttt{<REF\_B>} are reference tokens pointing to two lines introduced earlier in the sequence.

While geometric primitives also contain continuous parameters (coordinates, radii, angles, etc.), these parameters are not predicted by our model. Instead, they are treated as input to inform constraint generation. To incorporate this information, each primitive’s continuous parameters are embedded in a separate stream of tokens for input only. The generative process focuses on discrete constraints and dimensions that reference the primitives, leaving numeric values for dimensions to be resolved by the constraint solver. This design choice leverages the solver’s robust capacity to converge on valid parameter assignments, allowing the model to prioritize structural correctness and alignment with design objectives.

\subsection{SketchGraphs Dataset}
\label{appendix_sketchgraphs}
In addition to the main paper that describes how the SketchGraphs dataset was filtered and converted, we provide additional details regarding the motivation and practical considerations of each step are provided here. The primary goal of these refinements is to produce a clean, representative subset of sketches and ensure each example aligns with standard engineering constraints.

\subsubsection{Data Preprocessing}
In \Cref{tab:sketch_constraints_dimensions} we list out the supported constraint and dimension types in Onshape terminology that we included in the training data. Notably, we filter out less prevelant constraints (\texttt{Symmetric}, \texttt{Normal}, 
\texttt{Pattern}) and dimensions (\texttt{CenterLine}, 
\texttt{Projected}) to focus the learning task on the core geometric relationship types which form the backbone of sketch geometry. These filtered types represent higher-level constructs that can be equivalently modeled using more fundamental constraints and dimensions. For example, \texttt{Symmetric} constraints can be composed using a combination of \texttt{Midpoint}, \texttt{Equal}, and \texttt{Collinear} constraints. Similarly, \texttt{Pattern} constraints typically express repeated geometry with equal spacing, which can be reconstructed through a combination of \texttt{Equal} dimensions and manually replicated constraints. By removing these non-core constraints, we simplify the constraint vocabulary the model must learn while still covering the vast majority of design intent in sketches.

\begin{table}[ht]
    \centering
    \small
    \caption{Supported Constraints and Dimensions}
    \begin{tabular}{ll}
        \toprule
        \textbf{Constraints} & \textbf{Dimensions} \\
        \midrule
        \texttt{Coincident} & \texttt{Diameter} \\
        \texttt{Horizontal} & \texttt{Radius} \\
        \texttt{Vertical} & \texttt{Distance} \\
        \texttt{Parallel} & \texttt{Angle} \\
        \texttt{Perpendicular} & \texttt{Length} \\
        \texttt{Tangent} &  \\
        \texttt{Midpoint} &  \\
        \texttt{Equal} &  \\
        \texttt{Offset} &  \\
        \texttt{Concentric} &  \\
        \bottomrule
    \end{tabular}
    \label{tab:sketch_constraints_dimensions}
\end{table}

We next eliminate redundant constraints by deduplicating overlapping coincident points. We identify groups of points that all coincide and merge or remove duplicate coincident constraints among them. This deduplication of coincident points removes unnecessary edges in the constraint graph, reducing its complexity without altering the sketch’s geometry. This focuses the model on the unique geometric relationships and avoids penalizing it for not outputting repetitive constraints that do not add new information. To avoid bias from repeated structures, we also deduplicate very similar or identical sketches in the dataset. We detect and remove duplicate sketches so that each unique sketch structure is represented more evenly.
\begin{table}[h]
    \centering
    \small
    \caption{Statistics of the SketchGraphs dataset after preprocessing.}
    \begin{tabular}{lc|lc}
    \toprule
    \multicolumn{4}{c}{\textbf{Dataset-Level Statistics}} \\ \midrule
    
    Sketch Count & 2,784,964  & & \\
    \% FC        &  8.27    & \% Not Solvable  &  1.62 \\
    \% OC        & 16.11    & \% Stable (bins=4)   & 93.70 \\ \bottomrule
    \end{tabular}
    
    \vspace{1em}
    
    \begin{tabular}{lcccc}
    \toprule
    \multicolumn{5}{c}{\textbf{Sketch-Level Statistics}} \\ \midrule

    & \multicolumn{1}{c}{Mean {\scriptsize$\pm$Std}} 
    & \multicolumn{1}{c}{Min} 
    & \multicolumn{1}{c}{Median} 
    & \multicolumn{1}{c}{Max} \\
    \cmidrule(lr){2-5}
    Entity Count     & 14.68 {\scriptsize$\pm$7.27}  & 1    & 13    &  64 \\
    Constraint Count &  6.53 {\scriptsize$\pm$5.48}  & 0    &  5    &  52 \\
    Dimension Count  &  1.08 {\scriptsize$\pm$1.67}  & 0    &  0    &  42 \\
    \% Point FC      & 27.13 {\scriptsize$\pm$22.72} & 0.00 & 20.00 & 100.00 \\
    \% Curve FC      & 33.48 {\scriptsize$\pm$29.49} & 0.00 & 28.57 & 100.00 \\
    \bottomrule
    \end{tabular}

    \vspace{1em}

    \begin{tabular}{lcc}
    \toprule
    \multicolumn{3}{c}{\textbf{Constraint- and Dimension-Level Statistics}} \\ \midrule
    & \multicolumn{1}{c}{Type Frequency (\%)} 
    & \multicolumn{1}{c}{Sketch Frequency (\%)} \\
    \cmidrule(lr){2-3}
    Coincident         & 16.39\% & 31.13\% \\
    Horizontal         & 19.18\% & 64.20\% \\
    Vertical           & 11.16\% & 36.01\% \\
    Parallel           & 19.70\% & 42.26\% \\
    Perpendicular      & 13.56\% & 47.98\% \\
    Tangent            &  7.62\% & 13.47\% \\
    MidPoint           &  7.49\% & 20.79\% \\
    Equal              &  4.79\% & 13.20\% \\
    Concentric         &  0.11\% &  0.48\% \\
    \cmidrule(lr){1-3}
    Offset   & 43.15\% & 21.84\% \\
    Diameter & 48.21\% & 25.37\% \\
    Radius   &  5.98\% &  4.57\% \\
    Linear   &  2.66\% &  1.93\% \\
    Angle    &  0.01\% &  0.01\% \\
    \bottomrule
    \end{tabular}
    \label{tab:sketchgraphs-stats}
\end{table}

After applying the above filters, we verify each sketch’s constraints for solver solvability. Any sketch that the solver identifies as unsolvable is removed from the training set for the SFT model training. This step guarantees that the model trains only on valid, feasible sketches that correspond to a realizable geometry. We also exclude sketches that are grossly under-constrained, where the solver indicates many degrees of freedom remain, since they may not demonstrate clear constraint interactions for the model to learn. However, we add these sketches back for model fine-tuning.

Finally, we fix at least one point in each sketch to lock its position. Because the SketchGraphs data often provides no absolute anchor in the plane, many sketches exhibit degrees of freedom that allow global translation or rotation without altering constraints internally. In a typical design environment, at least one point or an entire component is fixed to serve as a reference. Fixing a point eliminates global translational and rotational degrees of freedom, effectively locking the sketch in a consistent pose.

\subsubsection{Processed Data Statistics}

Table~\ref{tab:sketchgraphs-stats} provides detailed statistics of the SketchGraphs dataset after preprocessing. At the dataset level, the resulting set contains approximately 2.8 million sketches. Among these, only 8.27\% of sketches are fully-constrained (FC), highlighting the rarity of sketches that require no additional constraints. Around 16.11\% are over-constrained (OC), while 1.62\% are unsolvable. A majority (93.70\%) of sketches are stable when stability is evaluated using a 4-bin discretization of geometry positions.

At the sketch level, the average sketch consists of about 15 geometric entities and contains roughly 7 constraints and 1 dimension, although there is considerable variation (standard deviation 7.27, 5.48, and 1.67, respectively). Additionally, point-level and curve-level fully-constrained percentages per sketch average at approximately 27\% and 33\%, respectively, indicating that most sketches are significantly under-constrained at the primitive level.

\Cref{tab:sketchgraphs-stats} also summarizes the distribution of geometric constraints and dimensions in the dataset. The \textit{Type Frequency} column reports the percentage of each constraint or dimension type relative to the total number of constraints or dimensions. The \textit{Sketch Frequency} column shows the percentage of sketches in which at least one instance of the constraint or dimension appears.

We observe that commonly used geometric constraints such as \texttt{Horizontal}, \texttt{Vertical}, \texttt{Parallel}, and \texttt{Coincident} dominate the dataset, consistent with standard sketching practices in parametric CAD modeling. More specialized constraints like \texttt{Concentric} or \texttt{Tangent} appear less frequently, which aligns with their more limited use in practice. 

For dimensions, \texttt{Diameter} and \texttt{Offset} are most frequent, as circular and offset features are prevalent in mechanical design. \texttt{Radius}, \texttt{Linear}, and especially \texttt{Angle} dimensions appear less often, consistent with their relatively specialized applications. These trends support the realism and representativeness of the dataset, suggesting it captures authentic usage patterns by human experts in professional CAD environments.

\section{Architecture and Experiment Details}
\setcounter{figure}{0}
\setcounter{table}{0}

We discuss additional details regarding the model architecture, training, and experiments.

\subsection{Experimental Setup}

All experiments are conducted on an AWS P5.48xlarge instance. The instance is equipped with eight NVIDIA H100 GPUs (80 GB HBM3 memory per GPU), 192 vCPUs, and 2 TB of system memory. 

A single epoch of RL training with the SketchGraphs dataset takes \(\sim \)3 days to train. This is primarily due to the frequent interactions with the CPU-based constraint solver and the fact that solve times can be highly varied. Roughly half of the training time is spent on GPU computation and half on detokenization and solver interaction. We expect custom optimizations could significantly reduce training time.

\subsection{Constraint-Level Accuracy Evaluation}
Evaluating constraint generation by direct constraint-level accuracy (i.e., exact matches between predicted and ground-truth constraints) is not meaningful for the constraint generation task. First, most sketches in the SketchGraphs dataset contain only a partial set of constraints defined by the original designer. Consequently, the ground-truth data does not necessarily represent the only valid or complete solution for fully constraining the sketch. Second, for a given geometric configuration, there often exist multiple valid constraint sets that can yield an equivalent, fully-constrained and stable sketch. For instance, the same geometry can be constrained either by a combination of horizontal and vertical constraints or by applying equivalent dimensional constraints, both of which are acceptable in practice. This makes exact constraint matching an unreliable indicator of functional correctness.

Instead, we evaluate generated constraint sets using functional metrics that better reflect real-world utility, as described in~\Cref{sec:problem}. These include whether the generated sketch is fully-constrained, stable, and solvable—metrics directly tied to the practical usability of the generated constraints in CAD workflows.

\subsection{Vitruvion}
\label{sec:appendix-vitruvion}
We use Vitruvion as the core constraint generation model for all post-training algorithms. Our implementation is adapted to work with the Fusion sketch representation, which treats all points as distinct geometric primitives. This differs from Onshape, which introduces the concept of sub-primitives -- geometric entities can own points (e.g., a line owns its start and end points). In the tokenized geometry sequence, each geometric entity is represented by its top-level primitive along with a nested list of its associated sub-primitives. The pointer network can then reference both sub-primitives and standard primitives within the index space of the tokenized geometry sequence. By contrast, in Fusion there is no concept of ``sub-primitives'' -- all indices in the tokenized geometry sequence are associated with independent primitives. When pre-processing the data, we combine duplicate points in the SketchGraphs data and initialize these as separate points (i.e. not owned by a curve).

We additionally include a learned embedding for each entity indicating whether or not the entity is fixed or not. As mentioned in \Cref{appendix_sketchgraphs}, at least one fixed entity is necessary to act as an anchor to the rest of the sketch. In order for an entity to be fully constrained, the constraint graph must connect to a fixed entity. We posit that this information is valuable for the task of fully constraining sketches.

Our implementation represents curves, circles, and arcs using 5 points extracted along the path of the shape. This differs from Vitruvion which uses the parameters of the shape such as start/end points, center, radius, and arc midpoint. Lastly, we model constraints using the given (user) order rather than ordering based on the referenced primitives.

Our model generates constraints and dimensions as a structured token sequence, where each token represents a geometric primitive, constraint type, or dimensional relationship. This sequence-based representation allows the model to flexibly express a wide variety of parametric relationships. With a proper detokenization step, these sequences can be converted into standard constraint and dimension definitions supported by commercial CAD tools. As a result, the generated outputs are not limited to a specific platform and can be directly imported into widely used software such as Fusion, AutoCAD, Onshape, and SolidWorks, enabling seamless integration with existing design workflows.

\subsection{Preference-based Optimization Algorithms}
\label{sec:appendix-PO}

The hyperparameters for our preference-based optimization algorithms are presented in Table \ref{tab:pref_hyperparam}. Both DPO and Expert Iteration (ExIt) methods are initialized from the SFT model and undergo 2 full rounds of data generation using a temperature of 1.0 followed by policy improvement. The DPO implementation has additional hyper-parameters: a $\beta$ parameter controls preference strength, a small SFT loss weight combines the DPO loss with a standard cross-entropy loss on the positive sample $\tau_w$, and a label smoothing weight reduces model overconfidence. These settings were determined through preliminary experiments to optimize model performance.

\begin{table}[ht]
    \centering
    \small
    \caption{Training hyperparameters for preference-based optimization algorithms.}
    \begin{tabular}{lrrr}
        \toprule
        \textbf{Hyperparameters} & \textbf{ExIt} & \textbf{DPO} \\
        \midrule
        Batch size                  & 64 & 64  \\
        Rounds (N)                  & 2 & 2  \\
        Learning rate               & 1e-6 & 1e-5 \\
        Sampling temperature (data) & 1.0 & 1.0 \\
        $\beta$ (DPO)               & - & 0.1 \\
        SFT weight                  & - & 0.05 \\
        Label smoothing weight      & - & 0.3 \\
        \bottomrule
    \end{tabular}
    \label{tab:pref_hyperparam}
\end{table}

In the data generation phase, ExIt uses rejection sampling to filter out any under-constrained, over-constrained, or unsolvable model outputs. For DPO, we find all pairs \((\tau_w, \tau_l)\) of model outputs for the same sketch where $\tau_w$ is fully-constrained and $\tau_l$ is under-constrained, over-constrained, or unsolvable. In order to help DPO better distinguish between the positive and negative examples, we limit $\tau_l$ to have less than 90\% fully constrained curves.

\subsection{RL algorithms}
\label{sec:appendix-RL}
For the rewards, we used $r_\text{unstable}=-0.25$ as a penalty for unstable sketches, $r_\text{NS}=-1.0$ as a penalty for not solvable sketches, $r_\text{OC}=-1.0$ as a penalty for over-constrained sketches, and $r_\text{F}=-0.5$ as a penalty for sketches resulting in other failures. Other training hyperparameter choices are shown in \Cref{tab:rl_hyperparam}.

\begin{table}[h]
    \centering
    \small
    \caption{Training hyperparameters for RL algorithms.}
    \begin{tabular}{lrrr}
        \toprule
        \textbf{Hyperparameters} & \textbf{ReMax} & \textbf{RLOO} & \textbf{GRPO}\\
        \midrule
        Batch size                  & 32 & 32 & 32 \\
        Group sample size           & - & 8 & 8 \\
        Learning rate               & 1e-5 & 1e-5 & 1e-5 \\
        Sampling temperature        & 1.0 & 1.0 & 1.0 \\
        Reference update timesteps  & 100 & 100 & 100 \\
        KL penalty added to rewards & 0.01 & 0.01 & 0.0 \\
        KL regularization $\beta$   & - & - & 0.01 \\
        Policy clipping threshold $\epsilon$   & - & - & 0.2 \\
        \bottomrule
    \end{tabular}
    \label{tab:rl_hyperparam}
\end{table}

\section{Additional results}
\setcounter{figure}{0}
\setcounter{table}{0}

\subsection{Diversity}

Table \ref{tab:diversity_results} presents the diversity metrics for constraint generation across different models. The Vitruvion base model demonstrates the highest diversity with 65.23\% unique generations and a relatively low Mean Intersection over Union (MIoU) of 0.623, indicating substantial variation between generated constraints. In contrast, RLOO and GRPO show the least diversity, with 32.11\% and 33.95\% unique sketches respectively, and high MIoU values exceeding 0.88, suggesting considerable overlap in their generations. Expert Iteration achieves a better balance, maintaining relatively high diversity (62.80\% unique) while improving on the base model's performance. Standard SFT and Iterative DPO fall between these extremes, with the latter showing moderately improved diversity metrics over SFT. 

\begin{table}[ht]
    \centering
    \small
    \caption{Diversity results computed across 8 generations per sketch. Unique@8 is the percentage of the time that the model generates a unique set of constraints for each sketch, compared to the other generations for the same sketch. We measure uniqueness with the Weisfeiler Lehman (WL) graph hash with 4 quantization bins. MIoU is the average intersection over union of the generated constraints between the other generations for each sketch.}
    \begin{tabular}{lcc}
        \toprule
        Model                & \% Unique@8 $\uparrow$ & MIoU@8 $\downarrow$ \\
        \midrule
        Vitruvion (base)     & \bf{65.23}  & \bf{0.623} \\
        SFT                  & 46.71 & 0.782  \\
        Iterative DPO        & 52.79 & 0.775   \\
        Expert Iteration     & 62.80 & 0.720   \\
        ReMax                & 35.80  & 0.877   \\
        RLOO                 & 32.11 & 0.892    \\
        GRPO                 & 33.95 & 0.881  \\
        \bottomrule
    \end{tabular}
    \label{tab:diversity_results}
\end{table}

\subsection{Number of DPO/ExIt Iterations}

\begin{figure}[ht]
    \centering
    \includegraphics[width=\linewidth]{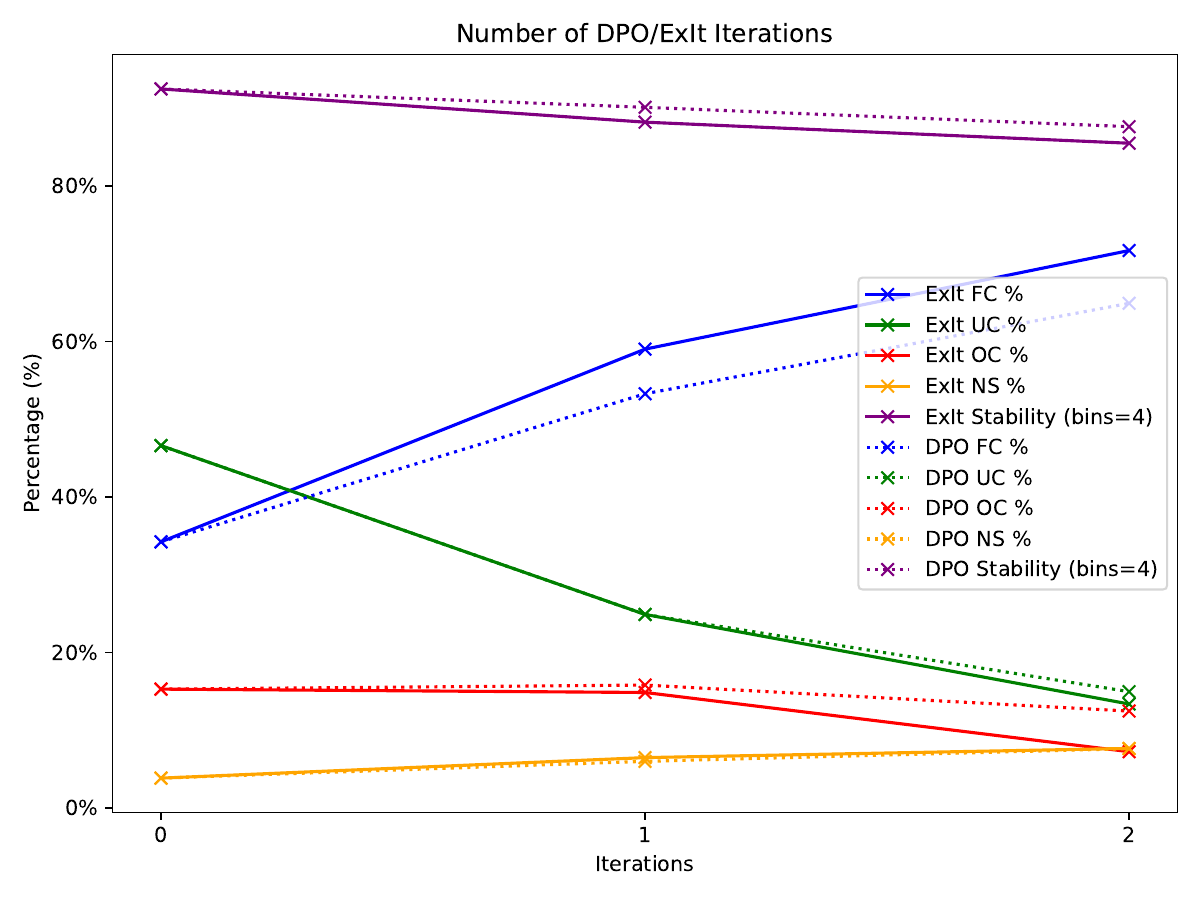}
    \caption{Performance across rounds for Iterative DPO and Expert Iteration. Results are the mean of $K=8$ samples. The initial model at $t=0$ is the SFT model}
    \label{fig:iterations_visual}
\end{figure}

\Cref{fig:iterations_visual} shows the performance of the preference-based optimization algorithms across training rounds. Expert iteration shows better performance at generating fully-constrained and not over-constraining sketches compared to DPO. One possible reason for this is that the process of selecting positive/negative example pairs for DPO is more restrictive since each positive (fully-constrained) example must be matched with an under-constrained or over-constrained example for the same sketch.

\subsection{RL Training curves}

\begin{figure}[ht]
    \centering
    \begin{subfigure}{\linewidth}
    \includegraphics[width=\linewidth]{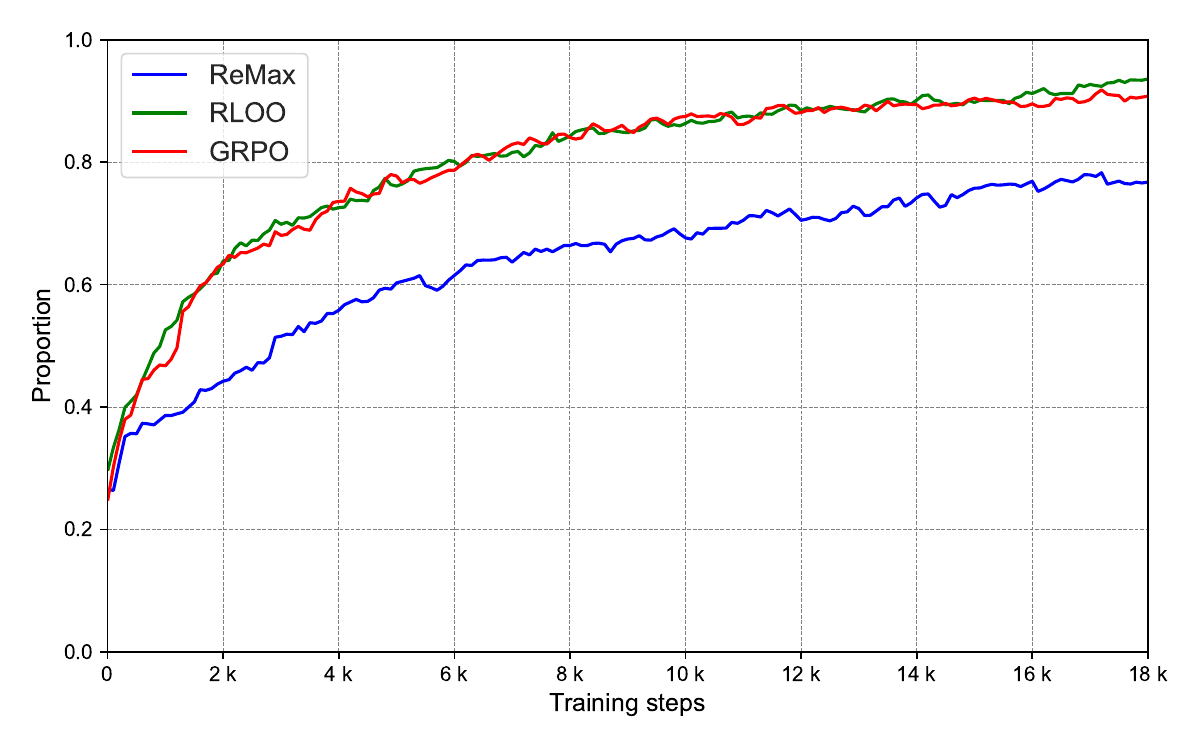}
    \caption{Proportion of successfully constrained sketches}
    \end{subfigure}
    \begin{subfigure}{\linewidth}   \includegraphics[width=\linewidth]{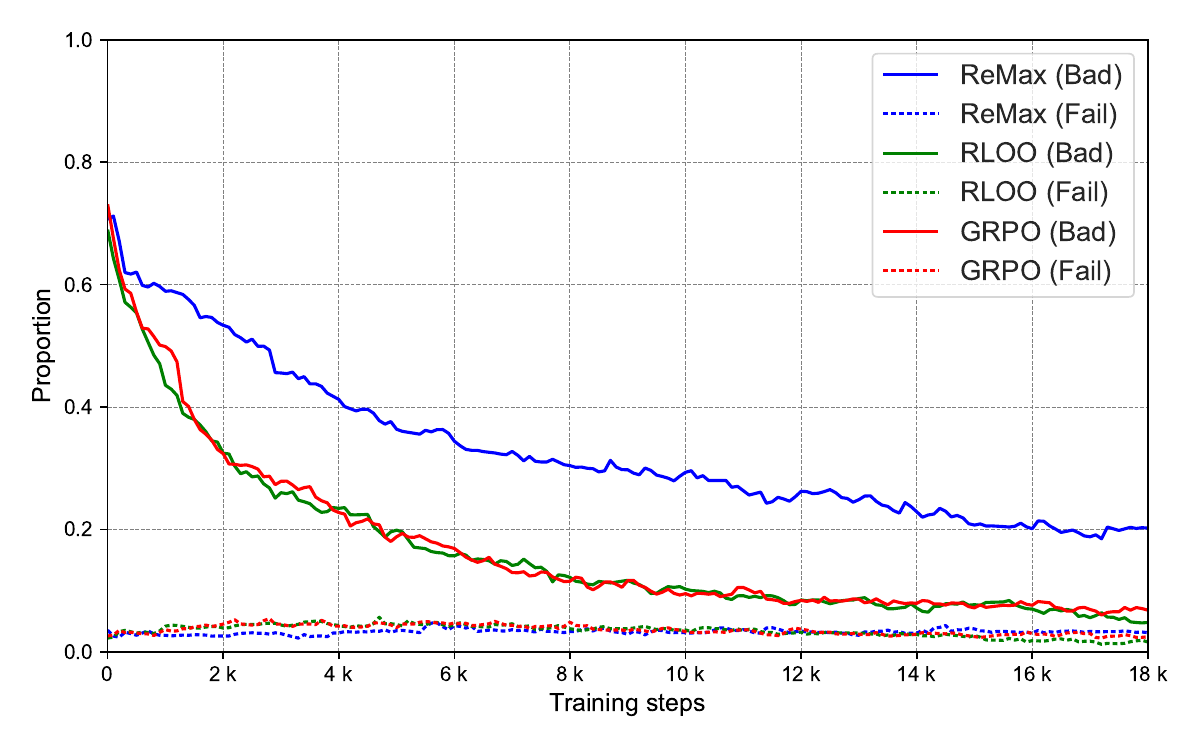}
    \caption{Proportion of unsuccessfully constrained sketches}
    \end{subfigure}
    \caption{Proportion of (a) successful sketches — defined as fully constrained but not over-constrained — and (b) badly constrained sketches which include under-constrained, over-constrained, or constraint solver errors, over the course of training for the RL methods ReMax, RLOO, and GRPO. Note that stability is not considered when determining whether a sketch is successful.}
    \label{fig:rl_train_visual}
\end{figure}

\Cref{fig:rl_train_visual} shows training performance over time for the online reinforcement learning algorithms.

\subsection{Impact of Sampling Parameters}
To assess the robustness of our approach with respect to sampling strategies, we conducted additional experiments varying the temperature $T$ and applying nucleus sampling with different top-$p$ values. Results are reported in~\Cref{tab:t_p_parameters}. We observe that increasing $T$ generally leads to more diverse constraint sequences, occasionally improving fully-constrained (FC) rates when combined with alignment methods such as RLOO and GRPO. Similarly, moderate nucleus sampling ($p = 1.0$) provides a favorable balance between exploration and reliability, whereas more aggressive truncation ($p = 0.5$) reduces diversity and causes the model to overfit to frequent constraint patterns, lowering FC performance. These findings indicate that alignment gains are robust within a reasonable range of sampling parameters, but extreme sampling settings can bias the generation toward either conservative or overly exploratory behaviors.

While all alignment methods benefit modestly from higher $T$ or larger $p$, the relative ranking of methods remains consistent. RLOO and GRPO show the least sensitivity, maintaining stable performance across all sampling settings, which suggests that their learned policies generalize better to variations in generation stochasticity. In contrast, SFT and other preference-based methods exhibit larger variance, indicating higher dependence on sampling choices.

\begin{table}[t]
    \centering
    \small
    \caption{Pass@4 results: $T$ refers to temperature and $p$ refers to the cumulative probability threshold in top-p sampling. }
    \begin{tabular}{lcccc}
        \toprule
        \multirow{2}{*}{Model} & \multicolumn{2}{c}{$T$ = 1.0} & \multicolumn{2}{c}{$T$ = 1.5} \\
        \cmidrule(lr){2-3} \cmidrule(lr){4-5}
                               & $p$ = 0.5 & $p$ = 1.0 & $p$ = 0.5 & $p$ = 1.0 \\
        \midrule
        Vitruvion (base)       & 13.69 &  15.94 &  12.96  & 14.31 \\
        SFT                    & 35.13 &  40.04  & 35.92  & 40.78 \\
        Iterative DPO          & 61.64  &  67.03  &  62.72  &  67.77 \\
        Expert Iteration       & 67.54  &  70.01  &  68.01  &  71.28 \\
        ReMax                  & 63.40  &  66.35  &  63.70  &  67.09 \\
        RLOO                   & \textbf{83.50}  &  \textbf{84.48}  &  \textbf{83.75}   & \textbf{84.89} \\
        GRPO                   & 82.21  &  84.01  &  83.49   & 84.17 \\
        \bottomrule
    \end{tabular}
    \label{tab:t_p_parameters}
\end{table}

\subsection{Impact of Reward Function Components}

Our original reward function was designed to encourage fully-constrained and stable sketches by maximizing the FC ratio and minimizing geometric movement during constraint solving. While effective at guiding the model toward functionally valid outputs, this setup inadvertently introduced a loophole. The model learned to maximize reward by adding excessive dimensions to overconstrain the sketch geometry, thereby reducing movement and achieving a high FC ratio. However, this behavior undermines the principles of parametric design, where the goal is for dynamic modifications and efficient exploration of design variations.

To address this issue, we extend the reward function with two additional penalty terms. The first term penalizes the total number of constraints and dimensions added, normalized by the number of geometric entities in the sketch. This discourages overly complex constraint sets. The second term penalizes over-reliance on dimensions by minimizing the ratio of dimensions to the total number of generated constraints and dimensions, promoting behavior more aligned with human experts who prefer geometric constraints over dimensional locking.

We denote this modified model as \textbf{RLOO with reward penalty}. When trained using the same hyperparameters as described in Table 1, the model achieves a slightly higher FC ratio of 72.79\% compared to Expert Iteration (ExIt), though it exhibits slightly lower geometric stability at 82.83\%. On average, it generates 3.7 dimensions and 11.54 constraints per sketch. In contrast, the original RLOO model without the new reward penalties produced an average of 19.5 dimensions and only 6.7 constraints, highlighting the effectiveness of the reward components in guiding the model away from degenerate solutions and toward more semantically meaningful constraint configurations.

\subsection{Human Evaluation Study}

To validate that our alignment methods produce constraint sequences that better align with human design intent, we conducted a human evaluation study with professional CAD designers. We designed a forced-choice perceptual study to compare constraint generation quality across five model variants: SFT (supervised fine-tuning), DPO (Direct Preference Optimization), ExIt (Expert Iteration), RLOO (REINFORCE Leave-One-Out), and RLOO with reward penalty. For each pairwise comparison, participants were presented with two images containing the same sketch but with constraints generated by different model variants.
\begin{table*}[ht]
\centering
\small
\renewcommand{\arraystretch}{1.2}
\begin{tabular}{l|c|c|c|c|c}
          & SFT       & DPO       & ExIt      & RLOO      & RLOO (reward penalty) \\
\hline
\textbf{SFT}         & --        & 24.67\%   & 16.67\%   & 82.67\%   & 36.00\%   \\
\textbf{DPO}         & 75.33\%   & --        & 36.67\%   & 90.67\%   & 48.67\%   \\
\textbf{ExIt}        & 83.33\%   & 63.33\%   & --        & 94.00\%   & 53.33\%   \\
\textbf{RLOO}        & 17.33\%   & 9.33\%    & 6.00\%    & --        & 8.00\%    \\
\textbf{RLOO (reward penalty)} & 64.00\%   & 51.33\%   & 46.67\%   & 92.00\%   & --        \\
\end{tabular}
\caption{
Pairwise preference study results between models. Each cell shows the percentage of times the row model was preferred over the column model (out of 150 comparisons per pair). Higher values indicate stronger relative preference.
}
\label{tab:user_study}
\end{table*}

\begin{figure} 
    \centering
    \includegraphics[width=\linewidth]{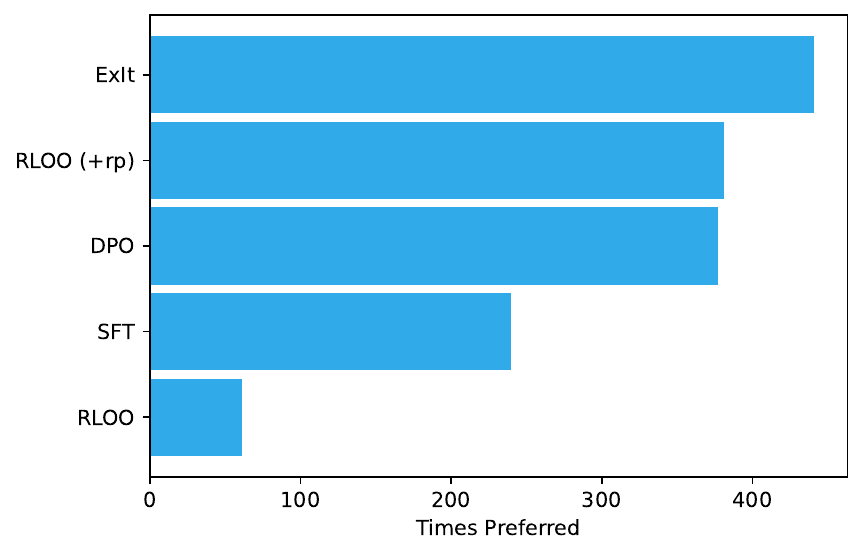}
    \caption{Number of times each model was preferred across 1500 pairwise comparisons by five expert designers. Each model appears 600 times as one of the two options to select.}
    \label{fig:human_preferences}
\end{figure}

The study included 30 representative sketches spanning different complexity levels, from simple rectangular profiles to more complex geometries involving arcs and tangent relationships, with each participant completing all possible pairwise comparisons between the five model variants, resulting in $\binom{5}{2} = 10$ comparison pairs per sketch. With 5 participants and 30 sketches, we collected 150 judgments per model pair, totaling 1500 pairwise comparisons.  The sketches were visualized with fully-constrained curves colored black and all other curves colored blue. Unstable sketches were purposefully removed in order to focus the participants on the quality of the generated constraints with respect to design intent.

\begin{figure*}
    \centering
    \includegraphics[width=\linewidth]{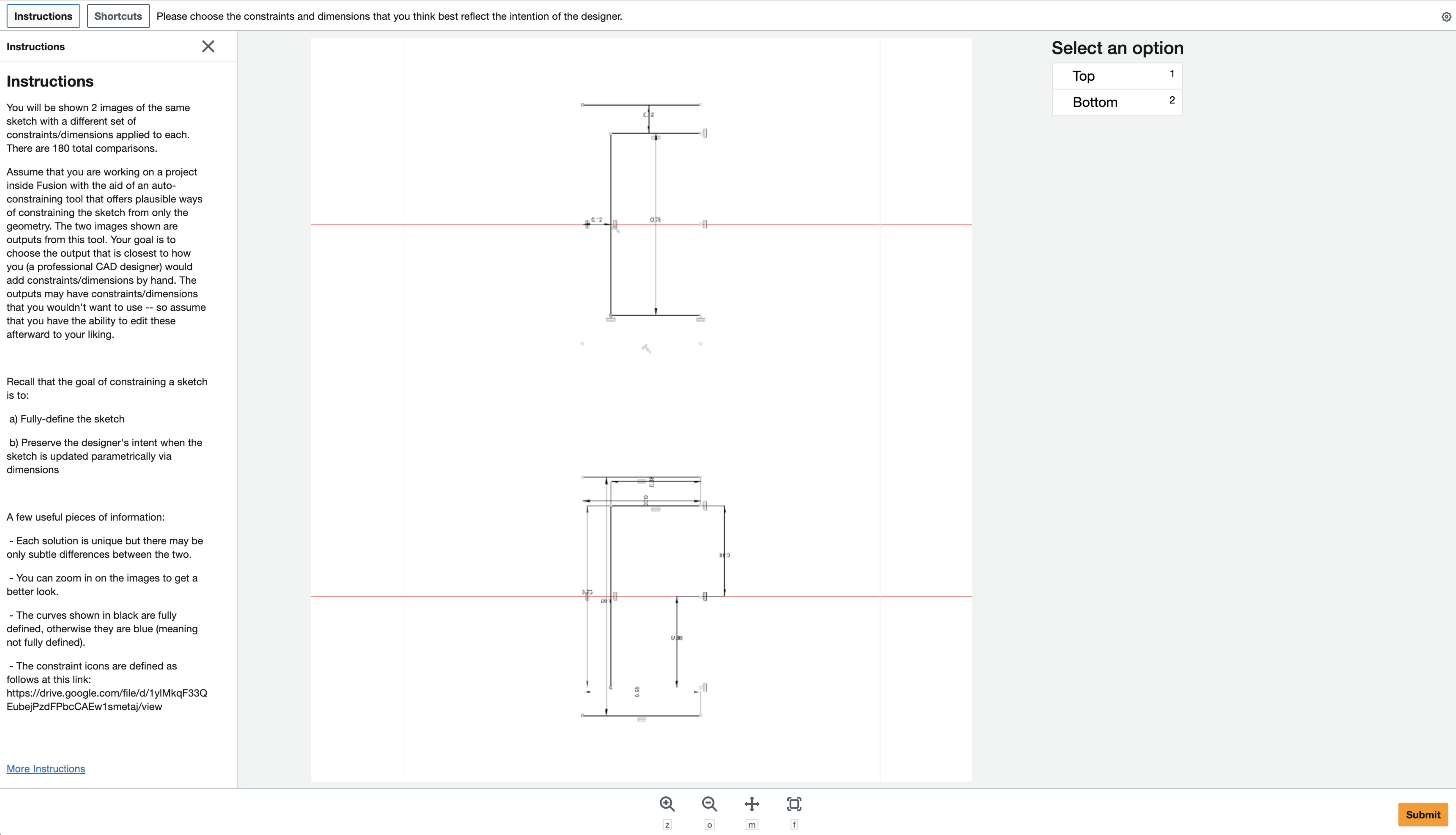}
    \caption{Screenshot of the user study. Participants were asked to follow the instructions in the left panel and choose the preferred sketch.}
    \label{fig:user_study}
\end{figure*}

A sample screenshot of what participants see while comparing sketches is shown in \Cref{fig:user_study}. Participants were asked to choose the set of constraints that they would use if tasked with constraining the sketch themselves and could make modifications on top of the generated constraints. 

\Cref{tab:user_study} presents the pairwise preference results, revealing a clear hierarchy: ExIt achieved strong performance, being preferred over SFT (83.33\%), DPO (63.33\%), and RLOO (94.00\%), while DPO outperformed SFT (75.33\%) and RLOO (90.67\%). Standard RLOO performed worst across all comparisons, with preference rates below 18\%. However, RLOO with reward penalty showed substantial improvement, being preferred over standard RLOO (92.00\% of the time) and achieving moderate performance against other methods. Compared to ExIt, we find that RLOO with reward penalty performs on par, with ExIt being preferred in 53\% of sketches on average. However, individual preferences vary: two participants preferred RLOO, two preferred ExIt, and one rated them equally. A similar trend is observed when comparing to DPO, where RLOO is preferred slightly more often on a sketch-by-sketch basis (48.67\% of the time), but a tie on individual preferences. These results suggest that the models are closely matched in overall performance, while reward design has a significant impact on the behavior of the RL model.

\Cref{fig:human_preferences} summarizes the total number of times each model was preferred by human evaluators in 1500 pairwise comparisons. Each model appears 600 times as a candidate in the evaluation. ExIt is overall the most favored model, reflecting its strong alignment with design intent. The vanilla RLOO model is least preferred due to its overuse of dimensions, which often reduces parametric flexibility. When reward penalties are added to RLOO to discourage unnecessary dimension use, its performance improves significantly, making it more competitive across designers.

\subsection{Failed Attempts}
Despite our efforts to leverage reinforcement learning for constraint generation, we encountered several dead ends. Each failed attempt underlines a fundamental challenge in aligning reward signals and exploration strategies with the requirements of geometric constraint generation. Below, we discuss three key failures, followed by brief summaries of the lessons drawn from each.

\subsubsection{PPO with a Learned Reward Model}
\label{subsec:appendix-pporm}
We first attempted to train a policy using PPO, guided by a learned reward model predicting how well the generated constraints would align with desired outcomes. This reward model serves as a surrogate model of the constraint solver, estimating the curve and point fully-constrained percentage, fully-constrained and under-constrained status, and stability. Unfortunately, the agent over-fit the reward model’s idiosyncrasies instead of genuinely improving constraint quality. In our case, PPO steadily increased the reward model’s score, but the rate of curve fully-constrained percentage actually dropped, which is evident that the policy was ``reward hacking'' the learned metric.

Several practical issues led to this failure. First, we lack diverse training samples for the reward model, especially for over-constrained or edge-case scenarios. The reward model was trained on two different settings, either on sparse per-sequence labels (only knowing the true evaluation metrics given an entire constraint set) or on per-constraint feedback. Both schemes suffered from limited coverage of failure modes. When PPO began producing novel constraint combinations outside the training distribution, the reward model was out of its depth. In our implementation, the reward model remained fixed during PPO fine-tuning; as the policy explored new regions of the constraint space, the frozen reward model’s prediction errors grew unchecked.

\subsubsection{PPO with Solver-based Rewards}
Another approach replaced the learned reward model with direct solver feedback, providing a reward only when the entire constraint sequence is generated. Although this feedback was unambiguously correct, it proved extremely sparse, the distribution of rewards remained highly skewed, with most episodes clustered near the lower or neutral end and only infrequent high-reward successes, causing training to collapse. For the policy gradient approach, such sporadic positive returns can still nudge the policy upward in proportion to the log probability of successful episodes. In contrast, the PPO algorithm sees little incremental feedback to guide learning, sudden high rewards are either clipped or overshadowed by large variance in advantage estimates.

\subsubsection{Logic-based Action Masking}
Finally, we tested logit masking to disallow certain ``invalid'' actions. In principle, this was meant to help by preventing the agent from exploring blatantly wrong moves. Surprisingly, this logit masking made learning worse for all our RL algorithms. One theoretical reason is that the mask, while eliminating invalid actions, also over-constrained the policy’s exploration. Contrary to expectations, blocking these actions harmed training. By never letting the agent attempt blatantly invalid moves, the model lost valuable negative feedback signals and drastically curtailed exploration. Another theoretical concern is that dynamic action masking can complicate the Markov Decision Process. So the issue is likely not that the concept of masking is invalid, but rather that it altered the learning dynamics in our specific setting.

\end{document}